\documentclass[10pt,conference]{IEEEtran}
\pdfoutput = 1

\usepackage{amsmath}
\usepackage{amsfonts}
\usepackage{amssymb}
\usepackage{amsthm}
\usepackage{nicefrac}

\usepackage{graphicx}
\usepackage{subcaption}
\usepackage{multirow}

\def\BibTeX{{\rm B\kern-.05em{\sc i\kern-.025em b}\kern-.08em
    T\kern-.1667em\lower.7ex\hbox{E}\kern-.125emX}}
    
\begin{document}

\title{Binary Diffing as a Network Alignment Problem\\ via Belief Propagation}

\author{
\IEEEauthorblockN{Elie Mengin}
\IEEEauthorblockA{SAMM, EA 4543 \\
Universit\'e Paris 1 Panth\'eon-Sorbonne, Paris, France \\
Quarkslab SA \\
13 rue Saint-Ambroise, Paris, France \\
elie.mengin@gmail.com}
\and
\IEEEauthorblockN{Fabrice Rossi}
\IEEEauthorblockA{CEREMADE, CNRS, UMR 7534 \\
Universit\'e Paris-Dauphine, PSL University, Paris, France \\
fabrice.rossi@dauphine.psl.eu}
}

\maketitle

\begin{abstract}
In this paper, we address the problem of finding a correspondence, or matching, between the functions of two programs in binary form, which is one of the most common task in binary diffing. We introduce a new formulation of this problem as a particular instance of a graph edit problem over the call graphs of the programs. In this formulation, the quality of a mapping is evaluated simultaneously with respect to both function content and call graph similarities. We show that this formulation is equivalent to a network alignment problem. We propose a solving strategy for this problem based on max-product belief propagation. Finally, we implement a prototype of our method, called QBinDiff, and propose an extensive evaluation which shows that our approach outperforms state of the art diffing tools. 
\end{abstract}

\begin{IEEEkeywords}
Binary Diffing, Binary Program Analysis, Graph Edit Distance, Network Alignment, Belief Propagation
\end{IEEEkeywords}

\section{Introduction}
Static program analysis is the process of analyzing and predicting the
possible execution behaviors and outcomes of a program without actually
executing it. It can be performed on the source code of the program or, with
more difficulty, on the binary executable. Static program analysis has a wide
variety of applications such as vulnerability detection, patch analysis,
malware detection, software clone detection, etc. 

In most of the cases, static analysis of binaries needs human expertise which
is leveraged using specific software tools. Among those tools, differs are
particularly useful as they allow the analyst to focus on the differences
between a previously analyzed program and the one currently under
investigation, enabling knowledge capitalization. Finding the differences
between two programs in binary form only is known as the \emph{binary diffing
problem}.

Several different formulations of the problem have been given, mostly depending
on the use case or the desired granularity but also implicitly induced by the
solving approach \cite{haq_survey_2021}. In this paper, we address the problem
of finding the best possible one-to-one correspondence between the respective
functions of two programs in binary form.

\begin{figure}
  \centering
  \includegraphics[width=\linewidth]{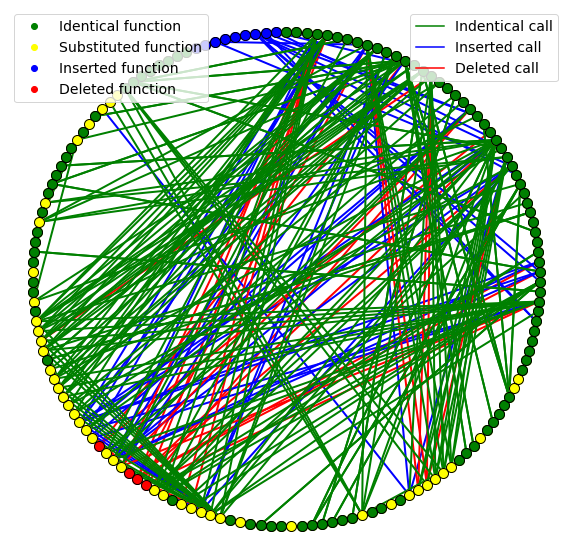}
   \caption{Binary diffing as an alignment of call graphs. This layout represents the superposition of functions (dots) and function calls (lines) of two binaries (libz-1.2.4.3 vs libz-1.2.6.1). Such mapping provides useful information to an analyst. In this figure, green dots and lines represent functions and calls that remained identical from a program to the other, and thus correspond to duplicated code. Blue (resp. red) elements represent inserted (resp. deleted) items, and may indicate added (resp. removed) functionalities. Yellow dots correspond to matched functions which content has been modified (substituted). They may thus record the functions that have been patched during the release. The figure highlights the interest of leveraging the function call consistency in order to find the best possible function correspondence.}
  \label{fig:graph-edit}
\end{figure}

Following previous authors \cite{hu_large-scale_2009, kostakis_improved_2011,
  bourquin_binslayer:_2013}, we leverage a graph edit formulation of binary
diffing: we find an (almost) optimal transformation of the call graph of
program $A$ into the call graph of program $B$, with respect to some specific
edit costs. We then show that this formulation is equivalent
to a network alignment problem. Following \cite{bayati_algorithms_2009}, we
propose an efficient approximate solver of this problem based on
\emph{max-product belief propagation}. In summary, our contributions are:
\begin{itemize}
	\item a new formulation of the binary diffing problem as a graph edit
          distance problem;
        \item an equivalent formulation as a network alignment problem;
        \item an efficient solver, QBinDiff, based on \emph{max-product belief propagation};
	\item a new diffing benchmark dataset consisting in more than 60
          binaries and over 800 manually extracted ground truth
          correspondences; 
	\item an extensive evaluation of our approach by comparing to other
          common matching methods, as well as other state of the art function
          similarity measures; 
\end{itemize}
Our experimental results show that the proposed approach outperforms other
diffing methods in almost all problem instances. Moreover, they highlight that
function similarity measures originally designed for near-duplicate detection
are not fitted to compute diffing assignments. Finally, they suggest that our
problem formulation is particularly adapted to address the binary diffing
problem.

The rest of paper is organized as follows. Section \ref{sec:binary-diffing} introduces
in more details the binary diffing problem and reviews some existing
solutions. Our proposed formalization as an optimization problem is described
in Section \ref{sec:formalization}, while Section
\ref{sec:framework} summarizes the maximization algorithm used to solve
the problem. Section \ref{sec:evaluation} is dedicated to the experimental
evaluation of our solution. 

\section{Binary diffing} \label{sec:binary-diffing}

\subsection{Graph representation and function matching}
In binary analysis, in order to consider its different potential
execution behaviors, a binary executable can be represented as a directed
attributed graph. In this graph, nodes stand for uninterrupted sequences of
instructions, called basic blocks, and edges indicate the possible jumps from a
basic block to another (conditional jumps, calls or returns). Such graph
theoretically represents all possible execution paths of the program. It is
known as the \emph{control-flow graph} (CFG) \cite{meng_binary_2016}. 

Another common representation of a program consists in a partition of the
control-flow graph according to the call procedures. The resulting directed
attributed graph is composed of nodes denoting the different program
functions and edges registering the calls among them. It is known as \emph{call
graph} (CG). This representation corresponds to a higher level of
abstraction than the CFG, closer to the developer point of view.

Note that retrieving both the CFG and the CG of a binary executable is a challenging problem that may not be solved exactly in some cases \cite{meng_binary_2016}. In this paper, we assume that both the CFG and CG can be reliably obtained from the executable programs.

In this paper, we define the binary diffing problem as the problem of \emph{matching} call graphs. We want to match functions from
one program to the functions of another such that they share similar functionalities (node content
similarity) and they call other functions in a similar way (induced edge
similarity). As a result, when a matching is satisfactory, the remaining differences
between the call graphs can be interpreted as meaningful modifications from
one program to the other.

As any binary diffing formulation, our definition requires a measure to assess the quality of a
matching between two call graphs. This measure
should evaluate the similarity of matched functions, as well as the relevance
of the resulting graph alignment. Therefore, any binary diffing instance is characterized by the given function similarity and topology similarity
measures. Once they are properly defined, we may formulate the binary diffing problem as an assignment problem which solution is the best one-to-one correspondence
between the functions of both program.

In the rest of this section, we present a short state-of-the-art of existing
methods to measure function similarity, as well as common proposed approaches
to compute the best function mapping. 

\subsection{Binary code similarity} \label{sec:binary-code-similarity}
The problem of measuring how much two pieces of binary code are similar is a fundamental problem of program analysis. Indeed, two seemingly different binary executables may have the exact same functionality. Such programs are said to be semantically equivalent while syntactically different. Conversely, two slightly divergent pieces of code may have very different behaviors when executed. Moreover, syntactic similarity is relatively easy
to compute but can lead to incorrect matching, while full semantic
characterization is undecidable (and heuristics tend to be computationally
expensive). Therefore, the definition of a similarity measure between two
binary functions generally involves an arbitrary trade-off between syntactic \cite{massarelli_investigating_2019, redmond_cross-architecture_2019, zhang_similarity_2020} or semantic comparisons \cite{cova_static_2006, gao_binhunt_2008, david_statistical_2016}.

Many recent approaches propose a mixed strategy. The idea is to use simple
syntactic features and to encode part of the function semantic through its
control-flow graph. For instance, 
Gemini \cite{xu_neural_2017}
introduces a \emph{Siamese graph neural network} to learn the common features of two
semantically similar functions. The model considers a very simple
representation of the function instructions as well as the basic block layout of the
CFG. It then embeds these features into a metric space where semantically
similar functions are likely to have close coordinates. Once every function
representation is projected into this metric space, pairwise similarity scores
can be computed very efficiently using common vector-based distance
computation routines.

Based on the same basis of Gemini, GraphMatching \cite{li_graph_2019} proposes
to enhance the model with an \emph{attention mechanism}
based on the structure of both function CFGs. However since it actively uses
the topology of both graphs during the similarity score computation itself, GraphMatching can not
benefit from fast vector-based distance computation as Gemini does. Therefore,
the time required to compute all pairwise similarity scores may rise
significantly with the size of the binaries.

Another alternative is Asm2Vec \cite{ding_asm2vec:_2019} which also provides
vector representations for binary functions but in an unsupervised way that
does not need matching pairs of functions. 
DeepBinDiff \cite{duan_deepbindiff_2020} improves over Asm2Vec by working at the
level of basic blocks. The embedding of a  basic
block is based on its content but also the one of its closer
neighbors. It uses an adaptation of a graph embedding algorithm,
\emph{text-attributed deep walk} (TADW) \cite{gao_deep_2018} to extract a vector
representation of each basic blocks among both binaries. To do so, it first
merges the \emph{inter-procedural} CFGs of both programs based on the binary symbols
and then runs the TADW algorithm to compute the embedding of each basic blocks
in this larger graph. Note that this approach is designed to proceed the
diffing at a basic block granularity, whereas ours seeks a mapping between the
functions of each binaries.

\subsection{Call graph matching}
Given a similarity measures between functions, one must now define a criteria of quality of the call graph alignment. In practice, such measure is closely related to the chosen function matching strategy.

The simplest solution for matching two call graphs consists in disregarding the call
graph structure itself and simply looking for a one-to-one mapping that maximizes the
sum of the similarities between the matched functions. In order to address the binary diffing problem, this would be the natural matching strategy used by methods originally designed to retrieve similar functions such as Gemini \cite{xu_neural_2017} or GraphMatching \cite{li_graph_2019}. Finding the best match reduces to an instance of the \emph{linear assignment problem} also know as the
\emph{maximum weight matching problem} (MWM). This is well known problem for
which optimal solutions can be found exactly in polynomial time, using
e.g. the \emph{Hungarian algorithm} \cite{kuhn_hungarian_1955}. The major
drawback of this approach is that the resulting mapping might be highly
inconsistent with regards to the call graph structure of the two programs.

To overcome this issue, other approaches such as BinDiff
\cite{dullien_graph-based_2005} and DeepBindiff \cite{duan_deepbindiff_2020}
propose instead to use matching algorithms designed to approximate the
\emph{maximum common edge subgraph problem} (MCS)
\cite{p._cordella_subgraph_2004}. Therefore, they implicitly define binary
diffing as an instance of the MCS. This problem consists in finding the node
correspondence which induces the maximum number of overlapping edges when
aligning the graphs \cite{bahiense_maximum_2012}. 
The general idea of those solutions is to expand in an iterative way the
partial solution by seeking potential matches in the neighborhood of the
current mapping (caller or callee of any already matched function).
Though in practice this strategy proved to provide
satisfying results, it suffers from a major limitation: by restricting new
matches to belong to the respective neighbors of the current partial mapping,
it prevents the assignment of potentially better non-local
correspondences. Therefore, this strategy mostly consists in finding a
locally-consistent mapping whereas a globally better assignment potentially 
exists.

\subsection{Graph edit distance}
A natural way to globally assess the quality of a matching is to consider it as a
particular case of \emph{graph edition}. One defines a set of graph edit
operations on both nodes and edges of the graphs, and assigns to them a
cost. The cost of a series of operations, also called an \emph{edit path}, is simply
the sum of the costs of said operations. Then, the edit path that transforms
graph $A$ into $B$ at the minimum cost is called an \emph{optimal edit path} and
the resulting edit cost is known as the \emph{graph edit distance}
\cite{riesen_structural_2016}. 

A matching can be viewed as a particular edit path in which matched nodes result
from an edition, whereas unmatched nodes in $A$ are
considered removed and unmatched nodes in $B$ inserted. The operation on the
edges (insertion/deletion/edition) are then completely induced by those on the
nodes (see Section \ref{sec:formalization} for details). Therefore, there is a
close relationship between an optimal matching and an optimal edit path.

Unfortunately, the computation of the graph edit distance of two arbitrary
graphs is known to be NP-complete and even APX-hard
\cite{lin_hardness_1994}. Though exact algorithms exist, they rapidly become
intractable as the number of vertices rises \cite{riesen_structural_2016}. In
practice, the computation of the GED of graphs of more than a hundred nodes
must be approximated. Note that the above-mentioned MCS problem is also
NP-complete. 

Several approaches previously proposed to compare programs in binary form
through a GED formulation \cite{hu_large-scale_2009, kostakis_improved_2011,
  bourquin_binslayer:_2013}. However, in order to compute an approximated
solution, all of them refer to Riesen and Bunke's linear programming
relaxation \cite{riesen_approximate_2009}, which reduces to a MWM formulation
of the binary diffing problem with a function similarity measure taking into
account the number of incident edge of each function. 

In this paper, we propose to directly address the GED problem through an
equivalent network alignment problem formulation. In this form, the
globally optimal edit-path can be efficiently approximated by means of a
message passing framework. 

\section{Formalization} \label{sec:formalization}

The novelty of our approach lies in the reformulation of the graph edit
distance calculation into a \emph{network alignment problem} (NAP) which can then be solved
(approximately) with a dedicated message passing algorithm. We give a formal
derivation of the NAP in the present section. 

\subsection{Binary diffing as a graph edit distance problem}
Let us consider two binary executables $A$ and $B$. We assume that adapted
disassembly tools are used to represent them by their respective call graph
$G_A = (V_A, E_A)$ and $G_B = (V_B, E_B)$. The vertices 
$V_A=\{1, \dots, n\}$ and $V_B=\{1', \dots, m'\}$ represent the functions of $A$ and
$B$. The edges $E_A=\{(i, j) | i, j \in V_A^2, i \neq j\}$ and
$E_B=\{(i', j') | i', j' \in V_B^2, i' \neq j'\}$ represent the function calls
(e.g. $(i,j)\in E_A$ encodes the fact that function $i$ calls function $j$ in
program $A$). Without loss of generality, self-loops (a.k.a. recursive calls)
are not taken into account (they can be accounted for at the level of the
function similarity calculation). 

We assume given two similarity measures. $\sigma_V$ measures the similarity
between two functions $i \in V_A$ and $i' \in V_B$ such that
$\sigma_V(i,i') = s_{ii'}$. $\sigma_E$ measures the similarity between
function calls. If $i$ calls $j$ in $A$ and $i'$ calls $j'$ in $B$, the
similarity of those calls is $\sigma_E((i, j), (i', j'))=s_{ii'jj'}$. We
assume that the similarities give values in $\left[0, 1\right]$. This enables
us to convert similarities into costs using $d_{ii'} = 1 - s_{ii'}$ and
$d_{ii'jj'} = 1 - s_{ii'jj'}$. Finally, we assume given two non-negative
constant values $d_\epsilon$ and $d_{\epsilon\epsilon}$ corresponding to the
cost of insertion or deletion of a function and call in a call graph.

\begin{table}
\centering
\caption{Graph edit operations and respective costs.}
\label{tab:GED}
\begin{tabular}{ll}
\hline
 Operation & Cost\\
\hline
 \emph{edit function} & $c(i \to i') = d_{ii'}$\\
 \emph{delete function} & $c(i \to \epsilon) = d_\epsilon$\\
 \emph{insert function} & $c(\epsilon \to i') = d_\epsilon$\\
 \hline
 \emph{edit call} & $c((i, j) \to (i',j')) = d_{ii'jj'}$\\
 \emph{delete call} & $c((i, j) \to \epsilon) = d_{\epsilon\epsilon}$\\
 \emph{insert call} & $c(\epsilon \to (i', j')) = d_{\epsilon\epsilon}$\\
\hline
\end{tabular}
\end{table}

We denote any series of graph edit operations $P = (op_1, \dots, op_k)$ an edit
path, and define $\mathcal{P}(A, B)$ as the set of all possible edit paths
that transform $G_A$ into $G_B$. Formally, if $(op_1, \dots, op_k)\in
\mathcal{P}(A, B)$, then $op_k(op_{k-1}(\ldots op_1(G_A)\ldots))=G_B$. We finally denote $C(P) = \sum_{i=1}^{k} c(op_i)$ the cost the edit path $P$.

Table \ref{tab:GED} lists the six possible graph edit operations we consider,
with their respective costs. In this paper, we restrict $\mathcal{P}(A, B)$ to
the set of \emph{restricted edit paths} \cite{bougleux_graph_2017}. An
important property of such paths is that they correspond to a unique mapping
between the functions of $A$ and those of $B$ (see the Appendix for
details). Note that unlike common GED definitions, our formulation implies a
constant cost for every function (or call) insertion or deletion, whatever its
content.

Based on these definitions, our formulation of the binary diffing problem consists in finding the minimal-cost edit path $P^*$ that transforms A into B. Formally:
\begin{equation}
	\tag{GED}
	\label{def:GED}
	\begin{aligned}
		P^* = & \underset{P \in \mathcal{P}(A, B)}{\arg\min} C(P),\\
		= &\underset{(op_1, \dots, op_k) \in \mathcal{P}(A, B)}{\arg\min} \sum_{i=1}^{k} c(op_i).
	\end{aligned}
\end{equation}

\subsection{Binary diffing as a network alignment problem}
We now reformulate our definition of the binary diffing problem as an equivalent instance of a \emph{network alignment problem}.

We first describe the diffing correspondences via a binary vector $\mathbf{x} \in
\{0,1\}^{|V_A| \times |V_B|}$ (where $|U|$ denotes the cardinality of the set
$U$) for which $x_{ii'}=1$ if and only if function $i$ in $A$ is matched with
function $i'$ in $B$. To ensure that each function from $A$ is matched to at
most one function in $B$ and vice versa, $\mathbf{x}$ must fulfil the following
constraints:
\begin{align}
\label{def:constraints}
\forall i \in V_A&, \sum_{j' \in V_B} x_{ij'} \le 1,& \forall i' \in V_B&, \sum_{j \in V_A} x_{ji'} \le 1.
\end{align}
A good matching should associate similar functions that have also similar
calling patterns. This can be captured in a cost matrix $Q \in
\mathbb{R}^{|V_A|^2 \times |V_B|^2}$ defined as follows:
\begin{equation*}
	Q_{ii'jj'} =
		\begin{cases}
		w_{ii'} & \text{ if } ii' = jj',\\
		w_{ii'jj'} & \text{ if }(i, j) \in E_A \text{ and } (i', j') \in E_B,\\
		0, & \text{ otherwise.}
	\end{cases}
\end{equation*}
with
\begin{align*}
  w_{ii'}&= s_{ii'} + 2 d_\epsilon - 1,&w_{ii'jj'} &= s_{ii'jj'} + 2 d_{\epsilon\epsilon} - 1.
\end{align*}
Using these definitions, it can be shown that computing the optimal edit path of \eqref{def:GED} is equivalent to solving the following network alignment problem:
\begin{equation}
	\tag{NAP}
	\label{def:NAP}
	\begin{aligned}
		\mathbf{x}^* =  & \underset{\mathbf{x}}{\arg\max}
		&&  \mathbf{x}^T Q \mathbf{x}\\
		& \text{subject to} && \forall i \in V_A, \sum_{j' \in V_B} x_{ij'} \le 1\\
		&&& \forall i' \in V_B, \sum_{j \in V_A} x_{ji'} \le 1
	\end{aligned}
\end{equation}
We provide a proof in the Appendix.

\subsection{Graph edit operation costs}

\subsubsection{Local vs global similarity trade-off}
The definition of the edit operation costs of any GED formulation usually
relies on carefully chosen data based considerations (see
e.g. \cite{bourquin_binslayer:_2013}). Costs have obviously an effect on the
quality of the matching but also on the ability of a solver to find an
approximately optimal solution. Moreover, because of
the difficulty of function comparisons, local similarities might be
inconsistent with the call patterns and there may be no solution optimal both
locally and globally. Therefore, a matching results from an inherent trade-off between local node similarity and global graph topology.

In order to control the trade-off one can decompose $Q$ into two terms and
weight them. We define $Q_1$ as the diagonal matrix in $\mathbb{R}^{|V_A|^2 \times |V_B|^2}$ with diagonal terms ${Q_1}_{ii'ii'}=w_{ii'}$ and $Q_2$ as
$Q_2 = Q - Q_1$. $Q_1$ gathers the function/node similarities while $Q_2$ contains all the
potential induced overlapping edges, called ``squares''. A potential square
consists of a pair of edges in both call graphs: $(i,j)\in E_A$ and $(i',j')\in
E_B$. If $i$ is matched to $i'$ and $j$ to $j'$, then the call structure is
preserved and, in a sense, forms a square (with two sides coming from the
matches and two sides coming from the calls).

Given a trade-off parameter $\alpha \in \left[0, 1\right]$, the objective
function of \eqref{def:NAP} can thus be modified into:
\begin{equation*}
	\alpha \mathbf{x}^T Q_1 \mathbf{x} + (1 - \alpha) \mathbf{x}^T Q_2 \mathbf{x}.
\end{equation*}
In terms of graph edit operations, this reformulation consists in
appropriately weighting the original edit operation costs. 

Notice that extreme cases for $\alpha$ correspond to some interesting
particular cases. When $\alpha=1$, we recover a maximum weight matching (MWM)
strategy which disregards the calls while $\alpha=0$ corresponds to a maximum
common edge subgraph instance (MCS) where function similarities are not
used. Therefore, our formulation can be seen as a balanced strategy
between the two most common binary code matching methods.

\subsubsection{Function content similarity}
In this paper, we propose a simple function similarity metric $\sigma_V$. It consists in a weighted Canberra distance \cite{lance_computer_1966} over the set of features given in Table \ref{tab:features}. During the computation, each feature is properly weighted according to its type. We distinguish content based (instructions), topological based (CFG layout), and neighborhood based features (CG callers and callees). Note that one of our feature refers to an instruction classification. This classification encodes each instruction using the class of its mnemonic and the ones of its potential operands. Our taxonomy consists in respectively 34 and 13 different mnemonic and operand classes.

Since most matching algorithms are sensitive to ties between function distances, we introduce a small perturbation to the resulting similarity scores. Assuming that the denomination of the functions is consistent with their order in terms of entry address, the similarity between function $i$ in $A$ and $i'$ in $B$ is being increased by the value $1 - \frac{|i - i'|}{\max(|V_A|, |V_B|)}$.

\begin{table}
\centering
\caption{Function features and respective weights used in our proposed similarity measure. The final similarity score is computed using the Canberra distance.}
\label{tab:features}
\begin{tabular}{lrl}
\hline
 Type & Weight & Features\\
\hline
 \multirow{3}{*}{Content} & \multirow{3}{*}{23} & total \# of instructions\\
  && \# of instructions per class\\
  && max \# of block instructions\\
 \hline
 \multirow{4}{*}{Topology} & \multirow{4}{*}{19} & \# of blocks \\
  && \# of jumps\\
  && max \# of block callers\\
  && max \# of block callees\\
 \hline
 \multirow{2}{*}{Neighboorhood} & \multirow{2}{*}{7} & \# of function callers\\
  && \# of function callees\\
\hline
\end{tabular}
\end{table}

\subsubsection{Function call similarity}
In order to measure the similarity of two function calls, we simply use a 0/1 indicator, i.e. $\sigma_E((i, j), (i',
j'))=1$ if and only if $(i, j)\in E_A$ and $(i', j')\in E_B$. Therefore, the matrix $Q_2$ can be computed through the Kronecker product of the affinity matrix of $G_A$ and $G_B$.

Finally, in order to compare with other state of the art methods, and because, in general, binary diffing favors recall over precision, we
set the insertion/deletion operation costs to $d\epsilon =
d_{\epsilon\epsilon} = \frac{1}{2}$. This forces the algorithm to produce a
complete mapping even when some assignments are of poor relevance.

As pointed out in Section \ref{sec:binary-code-similarity}, other similarity
measures have been proposed, some of them being much more sophisticated than
the one we propose to use. However, this paper aims at identifying the benefit of the proposed
matching approach and as such a simple metric seems more appropriate to
emphasize the effect of variations in matching. 

\section{Network alignment with max product belief propagation} \label{sec:framework}
The network alignment problem, sometimes also referred to as graph matching problem, is an important optimization problem that has been extensively studied for decades \cite{burkard_quadratic_1998}. Although it is not easier to solve the NAP than the GED, several efficient approximate algorithms have been proposed, based on spectral methods \cite{singh_global_2008, zhang_final:_2016}, convex or indefinite relaxations \cite{zaslavskiy_path_2009, lyzinski_graph_2016} or linearization \cite{klau_new_2009}.

In this section, we introduce a novel algorithm to efficiently approximate the
binary diffing problem as a network alignment problem. This algorithm is
inspired of a previous model of \cite{bayati_algorithms_2009}, which proposes
to address the integer program \eqref{def:NAP} through an equivalent graphical
model mode inference.

A graphical model is a way to represent a class of probability distributions
over some random variables \cite{koller_probabilistic_2009}. There is a strong
link between inference in graphical models and optimization, especially when
we consider Maximum A Posteriori (MAP) inference. This is the general problem
of finding the most probable value of some of the random variables given the
value of the rest of the variables. A particular case of MAP inference is
to find the mode of a probability distribution, i.e. the most probable value
of all its variables. One of the most efficient algorithm for MAP inference is
the \emph{max-product algorithm}. It consists in passing messages between the
vertices of the graphical model that represents the distribution (see
e.g. \cite{koller_probabilistic_2009}, chapter 13).

To leverage this algorithm, we design a graphical model that encodes both the
objective function and the constraints of NAP into a probability distribution
such that it assigns maximum probability to the optimal assignment of the 
NAP. Finding the mode of the distribution is then equivalent to solving the NAP.

Formally, matching vector $x$ is associated to binary random variables $X
=\{X_{ii'} \in \{0,1\}, ii' \in V_A \times V_B \}$. The constraints
\eqref{def:constraints} of NAP are encoded through Dirac measures $f_i: \{ 0,
1 \}^{|\partial f_i|} \rightarrow \{ 0, 1 \}$ and $g_{i'}: \{ 0, 1
\}^{|\partial g_{i'}|} \rightarrow \{ 0, 1 \} $ such that: 
\begin{align*}
\forall i \in V_A, f_i(x_{\partial f_i}) &=
	\begin{cases}
		1, & \text{ if } \sum_{j' \in V_B} x_{ij'} \leq 1,\\
		0, & \text{ otherwise.}
	\end{cases}\\
\forall i' \in V_B, g_{i'}(x_{\partial g_{i'}}) &=
  	\begin{cases}
    	1, & \text{ if } \sum_{j \in V_A} x_{ji'} \leq 1,\\
    	0, & \text{ otherwise,}
  	\end{cases}
\end{align*}
where $x_{\partial f_i} = \{x_{ij'} \in \mathbf{x}, j' \in V_B\}$, and similarly for $x_{\partial g_{i'}}$.

The probability distribution of the corresponding graphical model is then:
\begin{equation}
	p_X(\mathbf{x}) = \frac{1}{Z} \left[ \prod_{i=1}^n f_i(x_{\partial f_i}) \prod_{i'=1'}^{m'} g_{i'}(x_{\partial g_{i'}}) \right] e^{\mathbf{x}^T Q \mathbf{x}}
	\label{def:likelihood}
\end{equation}
Where the normalization constant $Z$ denotes the partition function of the model.

It is clear that the support of the distribution \eqref{def:likelihood} is equivalent to the set of feasible solutions in \eqref{def:NAP}. Furthermore, the mode of $p_X(\mathbf{x})$ corresponds to the optimal solution of \eqref{def:NAP}.

In our work, we introduce modifications to the original model of \cite{bayati_algorithms_2009} in order to speed up the computation and favor the messages convergence. Though the details and improvements of these modifications are out of scope for this paper, we provide a complete derivation of the message passing framework in the Appendix.

A key property of this model is the local structure of the message passing
scheme. This later limits the propagation of updates to the connected
components only, and therefore reduces the overall computation cost of the
problem when working on sparse graphs, which is generally the case of call
graphs. Moreover, it enables to discard some potential correspondences
considered too unlikely, and thus significantly reduce the size of the problem
solution set. This property is very useful to control the required computation
cost and memory usage of larger problem instance: as shown in
e.g. \cite{khan_multithreaded_2012}, the cost of one iteration of our algorithm is in $O(nnz(Q_1) + nnz(Q_2))$ where
$nnz(x)$ denotes the number of non-zero entries in $x$. Note that after the last iteration, we need to solve a MWM problem which adds a $O(nnz(Q_1)N + N^2 \log N)$ cost to the whole procedure, where $N = |V_A| + |V_B|$.

Our implementation includes a sparsity ratio parameter $\xi \in \left[0, 1\right]$ in order to remove a ratio of less probable correspondences and forces the algorithm to find a solution among the remaining ones.

\section{Evaluation}
\label{sec:evaluation}
This section is dedicated to a thorough evaluation of our proposed solution, named
QBinDiff, and to a comparison of its performances with a selection of
state-of-art diffing approaches. We describe first our evaluation
benchmark, then the chosen binary code similarity and finally the experiments. 

\subsection{Benchmark}
A diffing approach can be evaluated by comparing the mapping results with
``true'' assignments, known as the \emph{ground truth}. Unfortunately, such
assignments are not readily available and may be in fact very difficult to
determine in an objective way. As part of this work, we have built a
new benchmark that will be released to the research community.

\subsubsection{Benchmark design}
To select programs to include in the proposed benchmark, we have considered
several requirements. First, the source of the programs should be made readily available, within several different versions. This enables us to compile the program with symbols and thus ease the determination
of the ground truth. Second, well maintained source repositories with
explicit commit descriptions, detailed change logs, as well as a relatively
consistent function denomination over time are also very important features for the
ground truth extraction. Third, as this extraction is largely done manually,
program sizes should be ``reasonable''.

According to these considerations, we choose three well known open source
project to compose our experimental dataset, namely
Zlib\footnote{https://github.com/madler/zlib},
Libsodium\footnote{https://github.com/jedisct1/libsodium} and
OpenSSL\footnote{https://github.com/openssl/openssl}. Note that some of these
programs are amongst the most frequently used for evaluation in the
literature \cite{haq_survey_2021}.

For each of these projects, we first downloaded the official repository, then
we compiled the different available versions using GCC v7.5 for x86-64 target
architecture with -O3 optimization level and keeping the symbols. Once
extracted, each binary was stripped to remove all symbols, then disassembled
using IDA Pro v7.2\footnote{https://www.hex-rays.com/products/ida}, and
finally exported into a readable file with the help of
BinExport\footnote{https://github.com/google/binexport}. During the problem
statement, only plain text functions determined during the disassembly process
are considered. 

This extraction protocol provided us with respectively 18, 33 and 17 different
binary versions. For each project, given $n$ different versions of the program, we propose to evaluate our method in diffing all the $\frac{n(n-1)}{2}$ possible pairs of different executables. Statistics describing our evaluation dataset
are given in Table \ref{tab:dataset}.

\begin{table}
\centering
\caption{Description of our binary diffing dataset. The last six
  columns respectively record the number of different binary versions, the number of resulting diffing instances, the
  average number of functions and function calls and the average ratio of conserved functions in our manually extracted and extrapolated ground truth.}
\label{tab:dataset}
\begin{tabular}{lrrrrrrr}
\hline
 Program	& Vers. & Diff. & Nodes	& Edges	&  $\text{GT}$ & $\overline{\text{GT}}$ \\
\hline
 Zlib		& 18	& 153 & 153		& 235	& 0.99 & 0.96  \\
 Libsodium	& 33	& 528 & 589		& 701	& 0.98 & 0.79  \\
 OpenSSL	& 17	& 136 & 3473	& 18563	& 0.94 & 0.72 \\
\hline
\end{tabular}
\end{table}

Notice that both the average call graph size and density of the programs varies with the different projects. This variety will provide insights on the scalability of the diffing methods under
study as well as the effect of sparsity on our
solver.

\subsubsection{Ground Truth}
As recalled in Section \ref{sec:binary-diffing}, determining the 'true
assignment' between the functions of two given binaries is a difficult
task. In its strongest sense, this problem reduces to evaluate the semantic
equivalence between two pieces of code and is known to be undecidable
\cite{haq_survey_2021}.

In practice, when working on different versions of the same, open-source and
well documented binary, one may significantly reduce the difficulty by
carefully exploiting the human readable information available in both the
source code and the binary symbols. Moreover, some project repositories
include detailed commit descriptions that precisely record the modification
from a version to another. However, these information almost always refer to
the changes occurring during a release, and are thus only available for
contiguous versions. Therefore, in order to obtain the function assignment
among two arbitrary program versions, one must extrapolate the different
mappings of the releases that happened in the meantime.

Our ground truth extraction protocol has two steps. We first manually
determine what we think to be the function mapping that best describes the
modifications between two successive binary versions. This process is done
with regards to the Changelogs files, the source code and the unstripped
binaries. Excepted for few major project modifications, almost all the
functions are mapped from a version to its successor (see Table
\ref{tab:dataset}). 

Once all the contiguous version ground truth mappings are extracted, we deduce
all the pairwise diffing correspondences by extrapolating the mappings from
version to versions. Formally, if we encode the mapping between $A_1$ and
$A_2$ into a boolean matrix $M_{A_1 \to A_2}$ such that ${M_{A_1 \to
    A_2}}_{ii'} = 1$ if and only if function $i$ in $A_1$ is paired with
function $i'$ in $A_2$, then, our extrapolating scheme simply consists in
computing the diffing correspondence between $A_k$ and $A_n$ as follows:
$M_{A_k \to A_n} = \prod_{i=k}^{n-1} M_{A_i \to A_{i+1}}$. 

\subsection{Experimental setup}
All the experiments have been conducted on an identical
hardware\footnote{Intel Xeon E5-2630 v4 @2.20GHz}, using the
implementation provided by the authors when possible. 

Our method, QBinDiff, is used with its default parameters ($\alpha=0.75$ and
$\epsilon=0.5$), and within a maximum of $1000$ iterations. We set the
sparsity ratio parameter $\xi$ to $0$ for smaller projects Zlib and Libsodium and to
$0.9$ for OpenSSL.

We compare our method with BinDiff \cite{dullien_graph-based_2005}, a closed
source state-of-the-art binary diffing tool which uses a matching algorithm
very close to MCS, but based on different, non public, function similarity
heuristics.

In addition, our approach is compared to differs constructed by combinations
chosen among three function similarity measures and two matching algorithms
outlined in Section \ref{sec:binary-diffing}. All combinations are used. The
baselines are described below.

\subsubsection{Function content similarity}
To evaluate the impact of the function similarity measure on the diffing
process, three state-of-the-art binary code similarity approaches have been selected:
Gemini \cite{xu_neural_2017}, GraphMatching (GraphM.) \cite{li_graph_2019} and
DeepBinDiff (DeepBD.) \cite{duan_deepbindiff_2020}. 

Gemini and GraphMatching are supervised learning models that require to be
trained on multiple pairs of functions labeled as similar or different. As the
manual construction of such a dataset is tedious, existing methods usually
use a collection of functions extracted from slightly mutated programs, such
as different versions of an executable. Then, a pair of functions is
labeled as similar if they share the same (or very similar) name, and dissimilar
otherwise.

We applied this protocol to our dataset. Note that this should give a small competitive
advantage to differs based on Gemini and GraphMatching as their similarity measures
will be optimized on the specific type of functions found in the binaries
under study.

During the training process, we collected 85680 samples of 7276 differently named
functions from the unstripped binaries. 80\% of them were used as training
examples, 10\% as a validation set and the remaining 10\% were used to assess the final accuracy of the trained models. Both models were trained using their
recommended hyper-parameters. To compute the similarity score of two embedded vectors, Gemini uses a cosine similarity measure, whereas GraphMatching refers to a normalized euclidean metric. After the training, the models respectively
provided an estimated AUC\footnote{Area Under the ROC Curve} of $0.968$ and
$0.939$. 

We also trained DeepBinDiff instruction embedding model on each binary of our
dataset, following the protocol and the recommendations of the corresponding article
\cite{duan_deepbindiff_2020}. As DeepBinDiff provides embeddings of basic
blocks, we represent each function by the average of all its basic block
embeddings. 

\subsubsection{Matching}
Our matching algorithm is compared to the two most common methods found in the
literature, namely MWM and MCS,

The MWM matching strategy implicitly used by Gemini and GraphMatching consists
in solving the linear assignment problem based on the computed pairwise
similarity scores. The exact solution of this problem can be found using
conventional optimization solvers. 

Several algorithms have been proposed to approximate the MCS problem. In order
to compare with BinDiff and DeepBinDiff, we based our implementation on the
one used in DeepBinDiff. However, since a CG is usually much more dense than a ICFG, we limited the neighbor parameter
$k$ to $2$. Note that, to output a complete mapping, the algorithm terminates
by applying a MWM solver to the set of unmatched correspondences.

\subsection{Results}
The quality of a diffing result is measured using its precision and recall
with respect to the ground truth. We refer to the standard definitions of precision and recall i.e. $p = \frac{|M \cap G|}{|G|}$ and $r = \frac{|M \cap G|}{|M|}$ where $M$ and $G$ respectively correspond to the set of matched function pairs in the computed and ground truth assignments. Note that, except for BinDiff, all the
evaluated methods are designed to produce a complete mapping. In fact, none of
them includes a mechanism to limit the mapping of the most unlikely correspondences
during computation. Therefore, these matching strategies do not consider
precision but only focuses on recall. In future work, we will investigate the effect of rising the insertion/deletion operation costs $d\epsilon$ and $d_{\epsilon\epsilon}$ in order to favor the solution's precision
score.

\begin{figure}
  \centering
  \includegraphics[width=\linewidth]{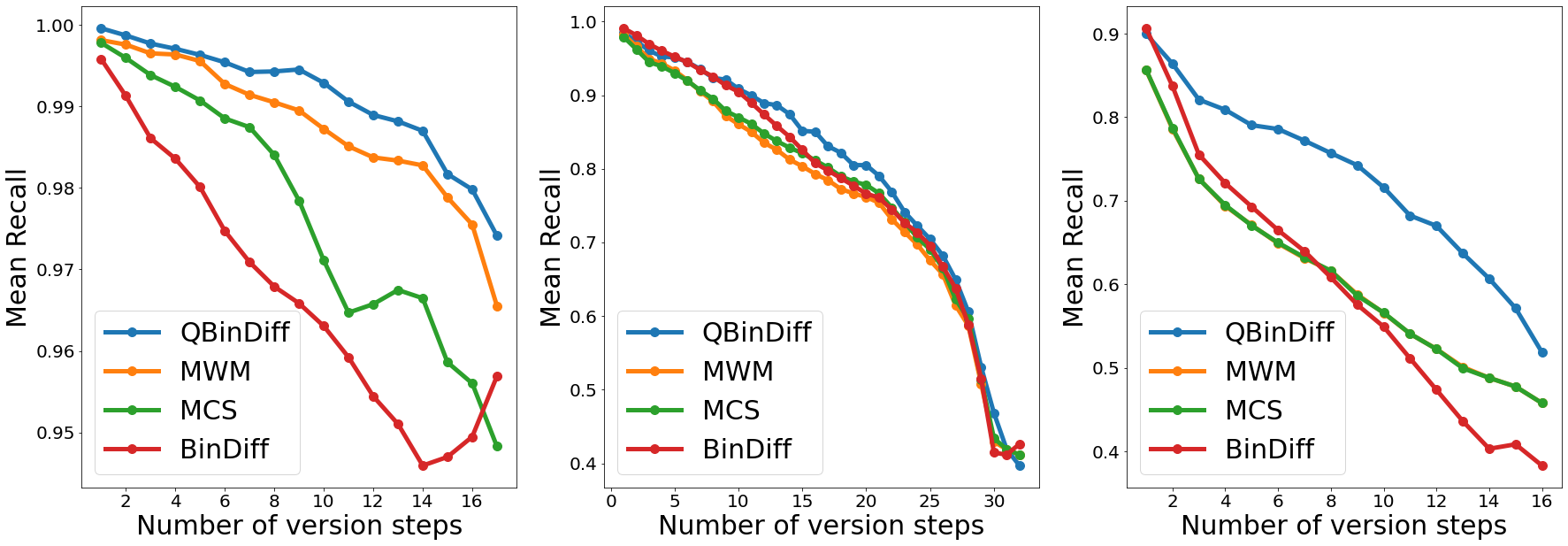}
   \caption{Average recall scores according to the program versions distance. Every matching method provides comparable near-optimal results while diffing very similar programs. As the distance increases, the performances of local matching strategies decline faster than our global approach.}
  \label{fig:acc-steps}
\end{figure}

Our experiments show that QBinDiff generally outperforms other matching
approaches in both precision and recall (see Table \ref{tab:acc}, QBinDiff+NAP
combinations). In fact, our method appears to perform clearly better at
diffing more different programs, whereas it provides comparable solutions on
similar binaries (see Figure \ref{fig:acc-steps}). This highlights that the
local greedy matching strategy of both MWM and MCS is able to provide good
solutions on simple cases but generalizes poorly on more difficult problem
instances. This results should be view as promising in the perspective of
diffing much more different binaries.

\begin{figure}
  \centering
  \begin{subfigure}[b]{0.49 \linewidth}
    \includegraphics[width=\linewidth]{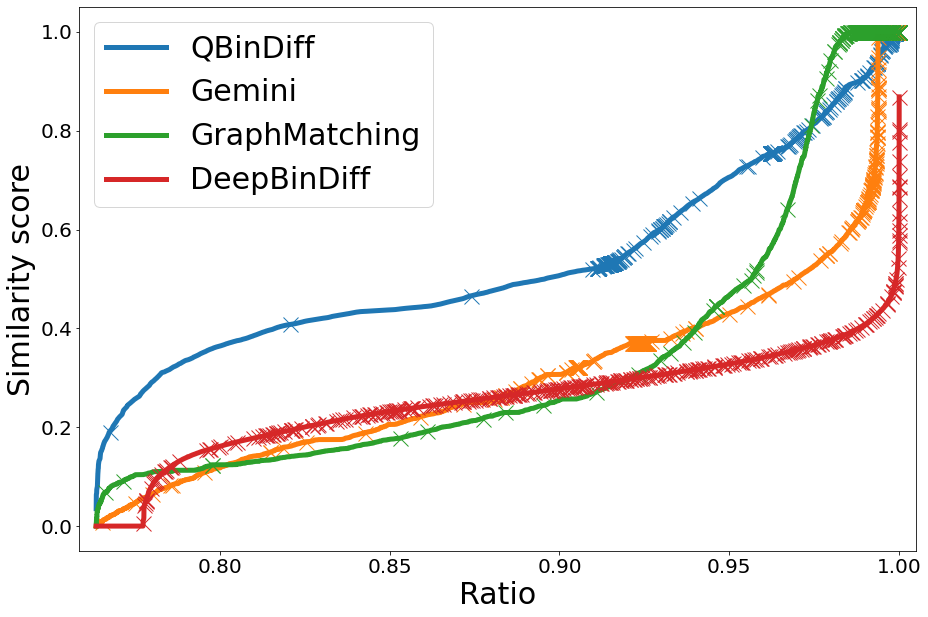}
    \caption{Pairwise similarity scores}
  \end{subfigure}
  \begin{subfigure}[b]{0.49	\linewidth}
    \includegraphics[width=\linewidth]{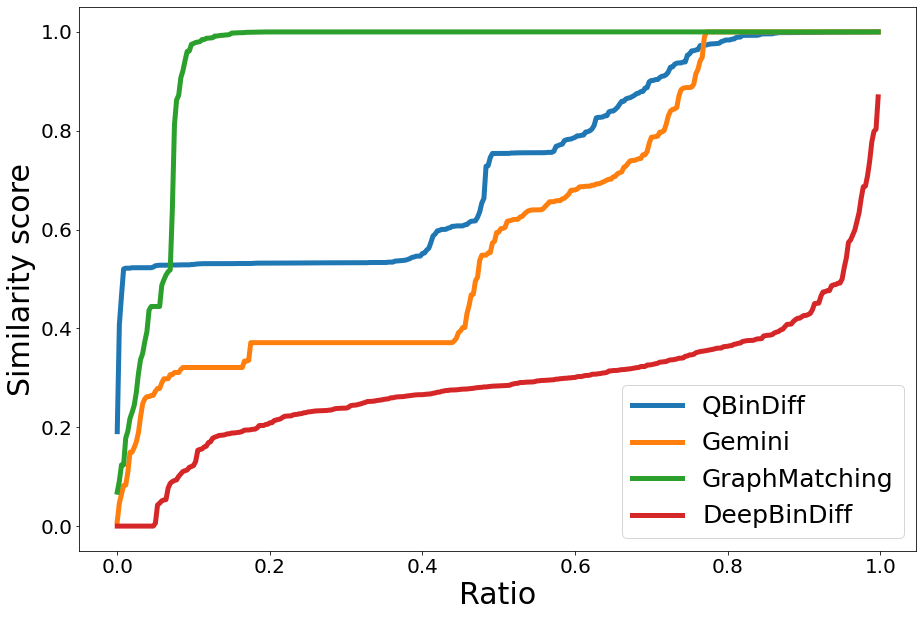}
    \caption{Ground truth similarity scores}
  \end{subfigure}
  \caption{Cumulative distribution function of all non-zero pairwise similarity scores (a) and of the ground truth pairs only (b) (libsodium-0.4.2 vs libsodium-1.0.3). The similarity scores in (a) that correspond to a ground truth correspondence are marked by a cross. GraphMatching appears to be well fitted to retrieve a large part of the correct matches but strongly deteriorates the score of some. QBinDiff provides a more balanced score but keeps almost all ground truth correspondence to a satisfying level.}
  \label{fig:sim-scores}
\end{figure}

Our NAP matching strategy can be applied with the state-of-the-art function
similarities chosen as reference. As observed with our custom metric, NAP
provides better assignments than other matching approaches. Moreover, it
appears that in almost all cases, the chosen matching strategy has more
influence than the similarity metric. More surprisingly, the use of these
complex models does not improve the accuracy of the resulting mapping, and
might even worsen it in some case. Since the topology of the graphs does not
change, this means that the computed similarity scores are not consistent with
the actual ground truth assignment. In fact, it appears that both Gemini and
GraphMatching models very accurately retrieve similar functions, but strongly
deteriorate the similarity scores of more diverging ones (see Figure
\ref{fig:sim-scores}). This is consistent with the original purpose of both
model and with the training dataset which labels as completely different two
similar functions with different names. In the case of DeepBinDiff, it seems 
that the scores of ground truth correspondences are distributed relatively
uniformly over the cumulative distribution function, which means that the
model itself does not provide sufficiently discriminative scores, and thus
leads to erroneous assignments. Moreover, we were not able to compute 
DeepBinDiff embeddings on OpenSSL binaries in reasonable time. Indeed, these
problem instances involve the factorization of the adjacency matrices of
graphs of over 100 000 nodes which is a very computationally intensive
task. 

\begin{table}
\centering
\caption{Average precision and recall scores for each combination of similarity
  measure (Similarity) and matching method (Matcher). The three tables correspond to the results on
  Zlib (top), Libsodium (middle) and OpenSSL (bottom) programs. The last two
  columns correspond to the similarity calculation time (Sim. calc.) and to
  the matching time (Time), both given in second.} 
\label{tab:acc}
\begin{tabular}{llrrrr}
\hline
 Similarity    & Matcher   &   Precision &   Recall &   Sim. calc. &   Time \\
\hline
 \multirow{3}{*}{QBinDiff}      & NAP       &       \textbf{0.955} &    \textbf{0.995} &          \multirow{3}{*}{3.0} &        0.2 \\
               & MWM       &       0.953 &    0.992 &              &        0.0 \\
               & MCS       &       0.946 &    0.985 &              &        0.0 \\
\hline
 \multirow{3}{*}{Gemini}        & NAP       &       0.953 &    0.992 &          \multirow{3}{*}{5.9} &        0.3 \\
               & MWM       &       0.936 &    0.974 &              &        0.0 \\
               & MCS       &       0.942 &    0.981 &              &        0.0 \\
\hline
 \multirow{3}{*}{GraphM.} & NAP       &       0.938 &    0.977 &         \multirow{3}{*}{77.1} &        0.8 \\
               & MWM       &       0.901 &    0.937 &              &        0.0 \\
               & MCS       &       0.927 &    0.964 &              &        0.0 \\
\hline
 \multirow{3}{*}{DeepBD.}  & NAP       &       0.909 &    0.946 &         \multirow{3}{*}{489.3} &        4.7 \\
               & MWM       &       0.820 &    0.853 &              &        0.1 \\
               & MCS       &       0.834 &    0.868 &              &        0.2 \\
\hline
 BinDiff       & BinDiff   &       0.943 &    0.975 &          0.3 &        0.9 \\
\hline
\\
\hline
 Similarity    & Matcher   &   Precision &   Recall &   Sim. calc. &   Matching \\
\hline
 \multirow{3}{*}{QBinDiff}      & NAP       &       0.722 &    \textbf{0.880} &         \multirow{3}{*}{13.3} &        6.6 \\
               & MWM       &       0.699 &    0.847 &              &        0.2 \\
               & MCS       &       0.704 &    0.854 &              &        0.3 \\
\hline
 \multirow{3}{*}{Gemini}        & NAP       &       0.714 &    0.863 &         \multirow{3}{*}{24.2} &        7.5 \\
               & MWM       &       0.668 &    0.802 &              &        0.2 \\
               & MCS       &       0.686 &    0.823 &              &        0.2 \\
\hline
 \multirow{3}{*}{GraphM.} & NAP       &       0.693 &    0.837 &        \multirow{3}{*}{293.9} &       21.7 \\
               & MWM       &       0.643 &    0.776 &              &        0.1 \\
               & MCS       &       0.670 &    0.806 &              &        0.3 \\
\hline
 \multirow{3}{*}{DeepBD.}  & NAP       &       0.664 &    0.796 &         \multirow{3}{*}{156.0} &       38.6 \\
               & MWM       &       0.585 &    0.702 &              &        1.9 \\
               & MCS       &       0.599 &    0.718 &              &        1.6 \\
\hline
 BinDiff       & BinDiff   &       \textbf{0.752} &    0.869 &          0.4 &        0.9 \\
\hline
\\
\hline
 Similarity    & Matcher   &   Precision &   Recall &   Sim. calc. &   Matching \\
\hline
 \multirow{3}{*}{QBinDiff}      & NAP       &       \textbf{0.605} &    \textbf{0.783} &         \multirow{3}{*}{88.6} &      213.3 \\
               & MWM       &       0.522 &    0.670 &              &       25.5 \\
               & MCS       &       0.522 &    0.670 &              &       24.4 \\
\hline
 \multirow{3}{*}{Gemini}        & NAP       &       0.577 &    0.685 &        \multirow{3}{*}{164.8} &      449.2 \\
               & MWM       &       0.400 &    0.467 &              &       24.9 \\
               & MCS       &       0.401 &    0.467 &              &       24.4 \\
\hline
 \multirow{3}{*}{GraphM.} & NAP       &       0.548 &    0.686 &      \multirow{3}{*}{36999.2} &     2187.5 \\
               & MWM       &       0.316 &    0.408 &              &       54.7 \\
               & MCS       &       0.317 &    0.409 &              &       55.1 \\
\hline
 BinDiff       & BinDiff   &       0.572 &    0.681 &          0.7 &        3.2 \\
\hline
\end{tabular}
\end{table}

\begin{figure}
  \centering
  \includegraphics[width=\linewidth]{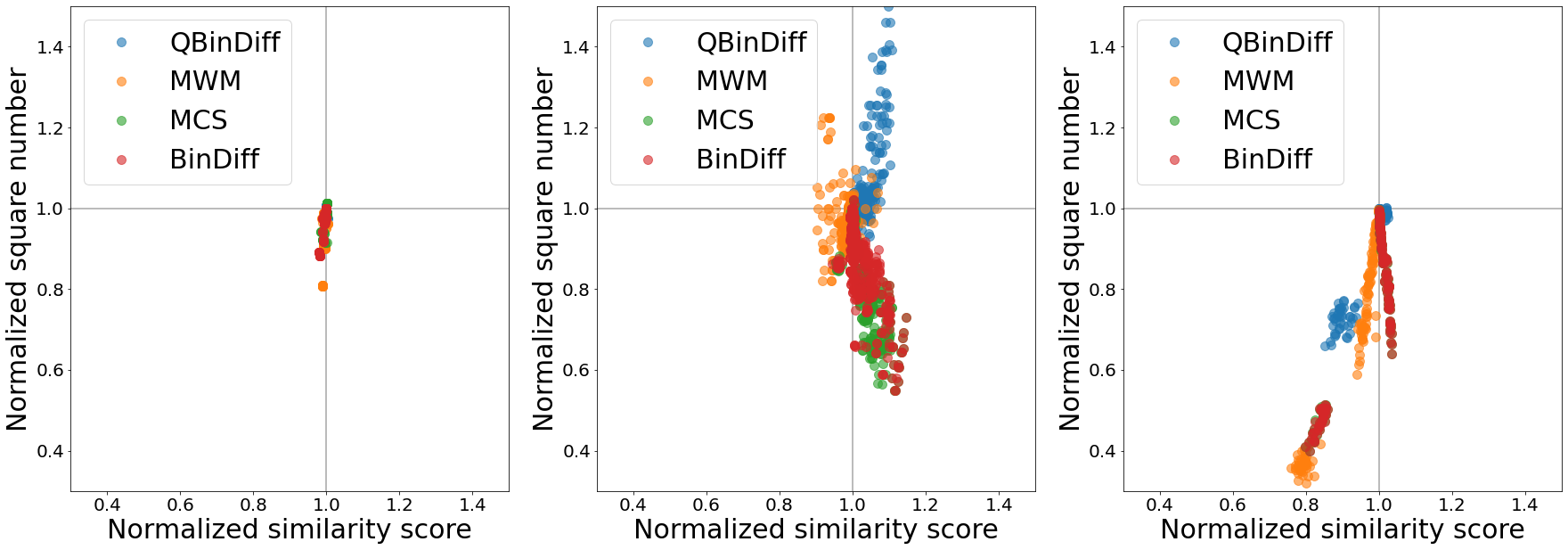}
   \caption{Relative similarity scores and square numbers of different matching methods compared to the optimal assignment. The grey lines record the normalized scores of the ground truth. For Zlib (left) and OpenSSL (right) binaries, the ground truth seems to be a near optimal NAP assignment in almost all cases. This result does not hold for Libsodium (center), as in some cases, assignments computed by QBinDiff are better in both function similarity and number of induced squares.}
  \label{fig:acc-ground-truth}
\end{figure}

An interesting analysis consists in comparing the different matching method
assignments to the ground truth correspondences in terms of function
similarity score and call graph alignment (see Figure
\ref{fig:acc-ground-truth}). It
appears that both Zlib and OpenSSL ground truth assignments are near-optimal
in both maximum weight matching and maximum common edge subgraph scores. This observation is consistent with our experimental results that
shows that a balanced network alignment matching strategy provide better
accuracy results than other approaches. More importantly, it justifies our
intuition that the proposed problem formulation as a network alignment problem
is very well suited to address the binary diffing problem. However, in some
cases, Libsodium correct assignments show to be sub-optimal in both function
similarity and graph topology. In these cases, the ground
truth mappings are inconsistent in both function content syntax and invoked
call procedures. We investigated these cases, and noticed that over the
versions, several functions were split in two such that a first trivial
function is solely designed to access a second core function actually
containing the whole function semantic. As we largely determined our ground
truth based on function names, we mapped full functions into their newly
created accessors. We discuss these specific cases in the next section.

Our experiments also suggest that our rather basic function similarity metric
provides scores that are consistent with the actual ground truth assignment
(see Figure \ref{fig:acc-ground-truth}). Moreover, on the contrary to
supervised learning models Gemini and GraphMatching, it produces less
discriminative scores. Though it might be view as a less informative metric,
it appears that this keeps the ground truth correspondences similarity scores
at a satisfying level and ultimately results in better solutions (see Figure
\ref{fig:sim-scores}). 

Finally, we recorded the computing time of each methods. As could be expected, it takes much
more time to approximate the NAP of two graphs than to compute the MWM or the
MCS. However, this can be controlled by raising the sparsity ratio parameter $\xi$,
at the cost of limiting the problem solution set and potentially resulting in
sub-optimal assignments. Moreover, it seems that better similarity scores speed-up the computation. This is due to the fact that the
algorithm finds more easily a satisfying local optima. Regarding the
processing times, it appears that, whereas the use of Gemini model does not
harm the required time, both GraphMatching and DeepBinDiff take very long time
to compute the pairwise similarity scores, which might be prohibitive for
larger programs.

\subsection{Limitations}
While it improves the state-of-the-art, our method could be further enhanced.  

A first limitation is that our approach is designed to find a one-to-one
correspondence between the functions of both programs. Thus, it can not
properly handle cases where a function in a binary is split into several ones
in the other program, or similarly, multiple functions are merged into a
single one. In such case, the information of both the function syntax and its
call graph relations is diluted into multiple chunks and may be harder to
retrieve. Note that, to our knowledge, this problem is common to all other
diffing methods, and that in practice, many function splits result in a core
function containing most of the semantic information, and few trivial
functions that are immediately called before or after it (as in the Libsodium
programs). Such schemes could be handled by a pre-processing step.

The other key property of our approach is that it is based on the assumption
that the true expected mapping is the optimal solution to the graph edit
distance problem. Although it is partially validated by our limited sized
experiments, there is no general available result that proves that this
intuition is verified in practice, especially for more complex cross-compiler or cross-architecture diffing instances. One may argue that this only depends on the
given graph edit operation costs definition. However, in practice, there is no
known function similarity metric that exactly encodes the functions semantic
and the interest of our method mostly rely on programs with rather similar
call graphs. Moreover, the trade-off parameter $\alpha$ that balances the node
and edge edit costs should be chosen carefully, which requires human expertise
and prior knowledge about the binaries under analysis.

Finally, an important drawback of our problem formulation is that it requires
a quartic memory matrix $Q$. Though we proposed to significantly reduce the
problem size by limiting the solution set to the most probable
correspondences, this relaxation inevitably induces information loss,
especially for large graphs where the relaxation must rise consequently. In
practice, binaries of several thousands of functions can be handled
efficiently. For larger programs, it might be better to first partition the
call graphs into smaller consistent subgraphs, and then proceed the matching
among them. Although this partition is not trivial and might result in
important diffing errors, it can be quite natural in modern programs for
example following its modules. 

\subsection{Threats to validity}

\subsubsection{Internal validity}
Our evaluation relies on a collection of diffing instances for which the ground truth assignment has been manually determined. Though we performed this extraction with regards to multiple sources of information such as source code, commit descriptions and unstripped symbols, we can not guarantee that our judgment is not biased, nor that it actually meets other experts opinion. Furthermore, any error or absence in our extracted mappings is later propagated in our extrapolation step. This may lower the confidence in the ground truth assignment between two distant versions. This threat is inherent to any manually determined assignments and can only be mitigated by releasing the dataset for the community to review.

\subsubsection{External validity}
Despite the relatively large number of proposed diffing instances, several factors still threaten the generalizability of our experiments. First, our benchmark only includes C programs taken from three open source projects. This is not representative of the variety of existing binaries. Moreover, all the executables were compiled with the same compiler, optimization level and targeted architecture. In future works, we will investigate the performance of our approach on programs built under different settings. Notice that this would probably require more sophisticated measures of similarity, able to efficiently handle greater syntactic differences. Last but not least, all our diffing instances compare different versions of a same program. Though the manual determination of an unanimous ground truth assignment between two different programs appears to be quite challenging, the evaluation of our method on such instances could be very instructive in the perspective of many applications such as the detection of vulnerability or of duplicate code.

\subsubsection{Construct validity}
The proposed comparison of our approach with other state of the art methods could also include threats to construct validity. First, all these methods are based on machine learning models that require a prior training step. We trained the models on the same dataset than the one we ran our experiments on. This could bias the resulting similarity scores, especially in case of overfitting. Moreover, we configured all models with their default parameters (recommended by the authors), though different settings could have provided better results. Finally, we must recall that none of the competitor methods where originally designed to address the exact same problem as ours. Indeed, both Gemini and GraphMatching have been initially proposed to retrieve near-duplicate functions whereas DeepBinDiff addresses the binary diffing problem at a basic bloc granularity.

\section{Conclusion}
In this paper, we introduced a new approach to address the binary diffing
problem. It is based on its reformulation as a graph edit distance
problem. This problem was shown to be equivalent to the network alignment
problem, for which we derived an approximate message-passing algorithm. We
proposed a new benchmark including hundreds of diffing ground truths and
used it to compare the proposed approach to state of the art binary diffing
methods.
	
Our experiments showed that our algorithm outperforms other existing
approaches in almost every problem instances. It also highlighted that the
matching strategy is a crucial part of the diffing process and has more
influence than the choice of the function similarity measure. Moreover, it
appeared that using similarity metrics originally designed to retrieve near
duplicate functions might actually harm the quality of the resulting
mapping. Finally, our results suggest that our problem formulation is a very
adapted way to address the binary diffing problem. 

Besides our formulation is quite natural and showed to result in more accurate
mappings, it also provides a proper metric for measuring program-wide
similarity. Indeed, any diffing assignment induces the (approximated) graph
edit distance between the two programs. Therefore, our approach could also be
used in a variety of metric-based analysis at a program level, such as library
retrieval, program lineage, etc. 

Finally, we believe that our graph matching algorithm could also be leveraged to perform diffing between matched functions in a post-processing step. This would results in a fined grained alignment between constitutive basic blocks of both functions and could provide to an analyst precious information about their exact differences.

\appendix

\subsection*{Proof of equivalence of \eqref{def:GED} and \eqref{def:NAP}} \label{sec:appendix}
In the following proof, we first show that the solution set of \eqref{def:GED} can be reduced to the one of \eqref{def:NAP}. Then we show that both objective functions are equivalent up to a sign and a constant term, which completes the proof.

Let $\mathcal{P}(A, B)$ be the set of all \emph{restricted edit path} transforming $G_A$ into $G_B$ \cite{bougleux_graph_2017}. This set consists in edit paths where any node can be removed only if its incident edges have been previously removed, and where any edge can be inserted only if its terminal nodes previously existed or have been inserted. Moreover, no nodes or edges can be successively inserted then edited, edited then deleted, inserted then deleted, or edited multiple times. Finally, overlapping edge must be considered as an edition and can not result from a deletion then an insertion.

It can be shown \cite{bougleux_graph_2017} that any edit path in $P \in
\mathcal{P}(A, B)$ can be fully characterized by a unique injective function
of a subset $\hat{V}_A$ of $V_A$ to $V_B$, and reciprocally. Such a mapping
can be encoded as a binary vector $\mathbf{x} \in \{0,1\}^{|V_A| \times
  |V_B|}$ such that $x_{ii'} =1$ if and only if $i \to i' \in P$ and 
satisfying the following constraints:
\begin{align*}
\forall i \in V_A&, \sum_{j' \in V_B} x_{ij'} \le 1,& \forall i' \in V_B&, \sum_{j \in V_A} x_{ji'} \le 1.
\end{align*}

Indeed, the injection implies that any node $i' \in V_B$ is the image to at most one node $i \in V_A$. Therefore, $\forall i' \in V_B, \sum_{j \in V_A} x_{ji'} \le 1$.
Moreover, any node $i \in \hat{V}_A$ has a unique image in $V_B$, so $\sum_{j' \in V_B} x_{ij'} = 1$, whereas any node $j \in V_A \setminus \hat{V}_A$ is not part of the injection and $\sum_{j' \in V_B} x_{jj'} = 0$.

Reciprocally, the constrained boolean vector defines a one-to-one mapping between the subsets of nodes $\hat{V}_A \in V_A$ and $\hat{V}_B \in V_B$. Thus, it implies a unique injection between $\hat{V}_A$ and the whole set of functions $V_B$.

Therefore, there is a bijection between the solution set of \eqref{def:GED} and the one of \eqref{def:NAP}.

Let us now evaluate the cost of any arbitrary edit path $P \in \mathcal{P}(A, B)$. Recall that this cost is completely induced by the edit operation on the nodes.

We first describe $C_V(P)$, the cost of the node operations in $P$. We distinguish the different possible operations such that:
\begin{equation*}
    \begin{aligned}
        C_V(P) = & \sum_{i \to i' \in P} c(i \to i') \\ & + \sum_{i \to \epsilon \in P} c(i \to \epsilon) + \sum_{\epsilon \to i' \in P} c(\epsilon \to i') \\
			   = & \sum_{i \to i' \in P} d_{ii'} - 2 d_\epsilon \\ & + \sum_{i \to \epsilon \in P} d_\epsilon + \sum_{i \to i' \in P} d_\epsilon + \sum_{\epsilon \to i' \in P} d_\epsilon + \sum_{i \to i' \in P} d_\epsilon \\
			   = & \sum_{i \to i' \in P} d_{ii'} - 2 d_\epsilon + \sum_{i \in V_A} d_\epsilon + \sum_{i' \in V_B} d_\epsilon \\
			   = & \sum_{i \to i' \in P} d_{ii'} - 2 d_\epsilon + |V_A| d_\epsilon + |V_B| d_\epsilon.
    \end{aligned}
\end{equation*}
In order to evaluate the cost of all the edges operations, we must consider the different possible configurations for pairs of nodes. But first, we must introduce the following notations:
$\delta^A_{ij} =
	\begin{cases}
	    = 1, & \text{ if } (i, j) \in E_A\\
	    = 0, & \text{ otherwise.}
	\end{cases}$, and similarly for $\delta^B_{i'j'}$.
	
We may now evaluate $C_E(P)$ such that:
\begin{equation*}
    \begin{aligned}
        C_E(P) =  & \sum_{i \to i' \in P} \sum_{j \to j' \in P} \Big[ d_{ii'jj'} \delta^A_{ij} \delta^B_{i'j'} \\ 
 & + d_{\epsilon\epsilon} \delta^A_{ij} (1 - \delta^B_{i'j'}) + d_{\epsilon\epsilon} (1 - \delta^A_{ij}) \delta^B_{i'j'} \Big] \\
        		& + \sum_{i \to i' \in P} \sum_{j \to \epsilon \in P} d_{\epsilon\epsilon} \delta^A_{ij}
        		 + \sum_{i \to i' \in P} \sum_{\epsilon \to j' \in P} d_{\epsilon\epsilon} \delta^B_{i'j'} \\     
        		& + \sum_{i \to \epsilon \in P} \sum_{j \to j' \in P} d_{\epsilon\epsilon} \delta^A_{ij}
        		 + \sum_{\epsilon \to i' \in P} \sum_{j \to j' \in P} d_{\epsilon\epsilon} \delta^B_{i'j'} \\
        		& + \sum_{i \to \epsilon \in P} \sum_{j \to \epsilon \in P} d_{\epsilon\epsilon} \delta^A_{ij}
        		 + \sum_{\epsilon \to i' \in P} \sum_{\epsilon \to j' \in P} d_{\epsilon\epsilon} \delta^B_{i'j'} \\        		 
			  = &  \sum_{i \to i' \in P} \sum_{j \to j' \in P} (d_{ii'jj'} - 2 d_{\epsilon\epsilon}) \delta^A_{ij} \delta^B_{i'j'} \\
				& + \sum_{i \in V_A} \sum_{j \in V_A} d_{\epsilon\epsilon} \delta^A_{ij} + \sum_{i' \in V_B} \sum_{j' \in V_B} d_{\epsilon\epsilon} \delta^B_{i'j'} \\
			  = &  \sum_{i \to i' \in P} \sum_{j \to j' \in P} (d_{ii'jj'} - 2 d_{\epsilon\epsilon}) \delta^A_{ij} \delta^B_{i'j'} \\
				& +  |E_A| d_{\epsilon\epsilon} + |E_B| d_{\epsilon\epsilon}. \\
    \end{aligned}
\end{equation*}

Putting all together, and denoting $C(P_0) = |V_A| d_\epsilon + |V_B| d_\epsilon + |E_A| d_{\epsilon\epsilon} + |E_B| d_{\epsilon\epsilon}$, the cost of any edit path $P$ is:
\begin{equation*}
    \begin{aligned}
    	C(P) = & C_V(P) + C_E(P) \\
        	 = & C(P_0) + \sum_{i \to i' \in P} d_{ii'} - 2 d_\epsilon \\ & + \sum_{i \to i' \in P} \sum_{j \to j' \in P} (d_{ii'jj'} - 2 d_{\epsilon\epsilon}) \delta^A_{ij} \delta^B_{i'j'} \\
        	= & C(P_0) + \sum_{ii' \in V_A \times V_B} x_{ii'} (d_{ii'} - 2 d_\epsilon) \\ +& \sum_{ii' \in V_A \times V_B} \sum_{jj' \in V_A \times V_B} x_{ii'} (d_{ii'jj'} - 2 d_{\epsilon\epsilon}) \delta^A_{ij} \delta^B_{i'j'} x_{jj'} \\
        	= & C(P_0) - \sum_{ii' \in V_A \times V_B} x_{ii'} w_{ii'} \\ & - \sum_{ii' \in V_A \times V_B} \sum_{jj' \in V_A \times V_B} x_{ii'} w_{ii'jj'} \delta^A_{ij} \delta^B_{i'j'} x_{jj'} \\
        = & C(P_0) - \sum_{ii' \in V_A \times V_B} \sum_{jj' \in V_A \times V_B} x_{ii'} w_{ii'jj'} x_{jj'} \\
        = & C(P_0) - \mathbf{x}^T Q \mathbf{x},
    \end{aligned}
\end{equation*}
where we simply use the fact that $d_{ii'} - 1 - 2 d_{\epsilon} + 1 = - w_{ii'}$ and similarly for $w_{ii'jj'}$.

Therefore, exploiting the one-to-one correspondence between an edit path $P$ and its boolean representation $\mathbf{x}$, we may turn the minimization of \eqref{def:GED} into a maximization problem and obtain \eqref{def:NAP}. $\blacksquare$

\subsection*{QBinDiff's message passing scheme}
We reproduce the messages derivation of \cite{bayati_algorithms_2009}, and introduce our modifications. For the sake of simplicity, we changed the message passing denomination of the original paper to best highlight their vectorial structure.

The messages $\mathbf{f} = \{f_{ii'}, w_{ii'} \neq 0\}$, $\mathbf{g} = \{g_{ii'}, w_{ii'} \neq 0\}$ and $\mathbf{h} = \{h_{ii'jj'}, w_{ii'} \neq 0 \cap w_{jj'} \neq 0\}$ are all initialized to $0$.

At each iteration, the algorithm computes the following updates:
\begin{align*}
	f_{ii'}^{(t+1)} = w_{ii'} & - \left( \max_{k \neq i} g_{ki'}^{(t)} \right)_+ - \gamma_{ii'}^{(t)} \\ & + \sum_{jj'}  \left[ w_{jj'ii'} + h_{jj'ii'}^{(t)} \right]_0^{w_{jj'ii'}} \\
	g_{ii'}^{(t+1)} = w_{ii'} & - \left( \max_{k' \neq i'} f_{ik'}^{(t)} \right)_+ - \phi_{ii'}^{(t)} \\ & + \sum_{jj'}  \left[ w_{jj'ii'} + h_{jj'ii'}^{(t)} \right]_0^{w_{jj'ii'}} \\
	h_{ii'jj'}^{(t+1)} = w_{ii'} & - \left( \max_{k' \neq i'} f_{ik'}^{(t)} \right)_+ - \phi_{ii'}^{(t)} \\ & - \left( \max_{k \neq i} g_{ki'}^{(t)} \right)_+ - \gamma_{ii'}^{(t)} \\ & + \sum_{kk' \neq jj'} \left[ w_{kk'ii'} + h_{kk'ii'}^{(t)} \right]_0^{w_{kk'ii'}},
\end{align*}
where: $x_+ = max(0, x)$, and $\left[ x \right]_a^b =
	\begin{cases}
	    = a, & \text{ if } x \leq a,\\
	    = x, & \text{ if } a < x < b\\
	    = b, & \text{ otherwise,}
	\end{cases}$
and where we introduced Bertsekas' $\epsilon$-complementary slackness mechanism \cite{bertsekas_auction_1992}:
\begin{align*}
\phi_{ii'}^{(t)} & =
	\begin{cases}
	    \epsilon, & \text{ if } f_{ii'}^{(t)} \neq \underset{k'}{\text{max }} f_{ik'}^{(t)}, \\
	    0, & \text{ otherwise.}\\
	\end{cases} \\
\gamma_{ii'}^{(t)} & =
	\begin{cases}
	    \epsilon, & \text{ if } g_{ii'}^{(t)} \neq \underset{k}{\text{max }} g_{ki'}^{(t)}, \\
	    0, & \text{ otherwise.}
	\end{cases}
\end{align*}

At the end of iteration $t$, the estimated mode is achieved by:
\begin{equation*}
	\hat{x}_{ii'}^{(t)} =
	\begin{cases}
	    = 1, & \text{ if } \hat{p}_{X_{ii'}}^{(t)} > 0,\\
	    = 0, & \text{ otherwise,}
	\end{cases}
\end{equation*}
where $\hat{p}_{X_{ii'}}^{(t)}$  denotes the log-ratio of the estimated marginal distribution of $x_{ii'}$:
\begin{align*}
	\hat{p}_{X_{ii'}}^{(t)} = w_{ii'} & - \left( \max_{k' \neq i'} f_{ik'}^{(t)} \right)_+ - \phi_{ii'}^{(t)} \\ & - \left( \max_{k \neq i} g_{ki'}^{(t)} \right)_+ - \gamma_{ii'}^{(t)} \\ & + \sum_{jj'} \left[ w_{jj'ii'} + h_{jj'ii'}^{(t)} \right]_0^{w_{jj'ii'}}
\end{align*}

\bibliographystyle{IEEEtran}
\bibliography{bibliography}

\end{document}